\documentclass{article}

\usepackage{microtype}
\usepackage{amsmath}
\usepackage{graphicx}
\usepackage{subfigure}
\usepackage{booktabs} 
\usepackage{caption}

\usepackage{hyperref}

\usepackage[accepted]{rldm2019}

\rldmtitlerunning{Bellman Error Minimization}

\begin{document}

\twocolumn[
\rldmtitle{Bellman Error Minimization \\
           Via Linear Programming Normalization}



\rldmsetsymbol{equal}{*}

\begin{rldmauthorlist}
\rldmauthor{Haining Yu}{to}
\end{rldmauthorlist}

\rldmaffiliation{to}{Independent Consultant}

\rldmcorrespondingauthor{Haining Yu}{hyu@bu.edu}

\rldmkeywords{Machine Learning, Dynamic Programming, Reinforcement Learning, Functional Approximation}

\vskip 0.3in
]



\printAffiliationsAndNotice{\rldmEqualContribution} 

\begin{abstract}
This paper proposes a new functional approximation approach to reduce Bellman error in high-dimensional dynamic programming and Reinforcement Learning problems. Using a classic dynamic programming problem (network capacity control in revenue management) as the motivational example, the paper illustrates that deep neural networks and linear programming approximation algorithms can be combined to derive approximate solutions to dynamic programming problems. Simulation results show the proposed approximation algorithms achieves competitive performance when compared with benchmark.
\end{abstract}

\section{Introduction}
Finding the optimal policy for high-dimensional dynamic programming problems remains the central challenge in reinforcement learning \cite{sb1998} and dynamic programming \cite{ber1995}. In high-dimensional problems, it is well known that DP suffers from the curse of dimensionality. Approximation algorithms is often the last resort.

When it comes to approximation, two broad classes of efforts exist. The first class attempts to find algorithms applicable to any DP problems by approximating the value function using a flexible (often nonparametric) family of functions (\cite{bt1996}; \cite{sb1998}; \cite{pow2007};\cite{mni2013}). The second class develops customized approximation based on industry and problem-specific structure; see examples like \cite{sdg2009} and \cite{tvr1998}.

Is there opportunity to unify the two approaches, that is, using the special structure of problem to reduce complexity, and taking full advantage the deep learning's ability to approximate complex nonlinear structure? We explore such opportunity in this paper. We are motivated by the following observations: on one hand, general approximation algorithms are often easy to implement, potentially reducing development time in practical applications. Meanwhile, a liberal interpretation to the no-free-lunch theorem \cite{wm1997} implies that performance of such universal algorithms can be limited by lack of customization. Customized algorithms, on the other hand, often can take advantage of specific problem structure and reduce the problem complexity. For specialized classes of problems, it is often possible to find customized approximation algorithms or even prove performance guarantee. At the same time, the custom-made nature of the solution often makes it difficult to be included as part of the machine learning pipeline. 

We perform a case study on the network capacity control problem \cite{tvr2004}. This classic dynamic programming problem makes it an ideal candidate to our study because it satisfies two criteria. First, its high-dimensional state space defies standard dynamic programming and justifies the need to pursue approximation. Secondly, its rich literature in approximation algorithms \cite{tvr1998} \cite{bp2003} makes benchmark possible. We propose a value function approximation that combines a deep neural network with customized linear programming (LP) formulation. We compare the performance of our approach with state-of-the-art approximation algorithms.
	
\section{Problem Setup}
We define the network capacity control problem following Chapter 3 of \cite{tvr2004}. A firm sells $N$ types of products. When every unit of product $n$ is sold, $n=1,...,N$, the firm collects revenue $r_n$. Selling a unit of product $n$ also consumes $a_{n,m}$ units of resource $m$, where $m = 1,..., M$. For simplicity we assume $a_{n,m} \in \left\{0,1\right\}$. We use vector $\textbf{a}_n = (a_{n,1}, ..., a_{n,M})$ and matrix $\textbf{A} = (a_{n,m})$ as shorthand when convenient.

Time is discretized into $T$ time steps, $t=0$ being the earliest. For any time step $t$, let $\textbf{p} = (p_1,..., p_N)$ be the probability vector for product $n = 1,...,N$ to arrive. We assume that the arrival processes for multiple products are independent, and that $\sum_{n=1,...,N}p_n \ll 1$ aka no two requests arrive in the same time step. At time $t$, let $x_m(t)$ be the remaining capacity of resource $m$ , $m=1,...,M$. The vector $\textbf{x}(t) = (x_1(t),...,x_M(t))$ captures the state of the network. If product $n$ is sold, the state of the network changes to $\textbf{x}(t)-\textbf{a}_n$.

The goal of the firm is to maximize total undiscounted reward (revenue) for the entire time horizon $T$. The central problem for the network capacity control problem is: at time $t$, given current resource capacity $\textbf{x}(t)$ and a request for product $n$, should the firm accept or reject the request? A policy $u(\textbf{x}(t),t,n)$ is a function that maps state $(\textbf{x}(t),t,n)$ to a binary decision of acception / rejection. We use $u = 1$ for acception and $u=0$ for rejection. For notational convenience, we will drop the time dimension in the following derivation.

Under optimal policy, the expected reward function $V_t(\textbf{x})$ satisfies the Bellman Equation:
\begin{equation}
\label{eqn_bellman1}
V_t(\textbf{x}) = E\left\{\max_{u \in \Psi (\textbf{x})}\left[ r_n u(\textbf{x},t,n) + V_{t+1}(\textbf{x}-\textbf{a}u) \right]\right\}
\end{equation}
with boundary condition 
\begin{equation}
\label{eqn_boundary}
V_{T}(\textbf{x}) = 0, \forall \textbf{x}
\end{equation}
In the above equation, $\Psi (\textbf{x})$ is the set of all valid policy functions. We assume there is no back-order, i.e., if $\textbf{x} \ge \textbf{a}_n$ is not true, $u(\textbf{x},t,n) = 0$. 

Given any value function $V_t(\textbf{x})$, the opportunity cost for any request for product $n$ is $V_t(\textbf{x})-V_t(\textbf{x}-\textbf{a}_n)$. The policy below then can be used to accept/reject any request:
\begin{equation}
\label{eqn_optimal_policy}
u(\textbf{x},t,n) = 1 \textrm{ iff }  \textbf{x} \ge \textbf{a}_n \textrm{ and } r_n \ge V_t(\textbf{x})-V_t(\textbf{x}-\textbf{a}_n)
\end{equation} Please note that equation (\ref{eqn_optimal_policy}) defines a policy given any value function. To emphasize this point, we can rewrite the above equation as $u(\textbf{x},t,n) = \Pi (V_t(\textbf{x}))$.

The network capacity control problem is of central importance to profitability of many  companies, including but not limited to most companies in airline and hospitality industries. Thus Problem (\ref{eqn_bellman1}) has been extensively studied; see \cite{tvr2004} for details. For a typical network with 10 resources, 100 units of initial capacity for initial capacity, and 1,000 time steps, the number of possible state and time combination is $10^{23}$, too large to calculate by backward induction. Currently, best approximation algorithms is based on decomposition approaches, e.g., in \cite{tvr1998} and \cite{bp2003}. In such algorithms, the dynamic programming problem in high-dimensional space is split into multiple (often independent) one-dimensional resource control problem. The latter set of problems can be solved by backward induction.

\section{Policy Generation}
\subsection{Functional Approximation and Linear Programming Normalization}
To generate a good policy $u^*(\textbf{x},t,n)$, \cite{bt1996} proposes approximating the value function $V_t(\textbf{x})$ with $V_t^*(\textbf{x})$, an approximate value function defined by a deep neural network. In theory, it is well known that a ``big enough" neural network can capture complex nonlinearity for a large family of continuous functions. In reality, however, a vanilla neural network can lead to poor performance and long development time. 

We define a LP-normalized functional approximator as:

\begin{equation}
\label{eqn_lp_normalizer}
V_t^*(\textbf{x}|\theta) = U_t(\textbf{x})W_t(\textbf{x}|\theta)
\end{equation}

$V$ is the product of two functions, $W$ and $U$. $W_t(\textbf{x}|\theta)$ is the output of a neural network. $\textbf{x}$ and $t$ are its input; $\theta$ are weights and biases. Independent from $\theta$, $U_t(\textbf{x})$ is the solution to the LP:

\begin{eqnarray*}
U_t(\textbf{x}) = && \max \textbf{r}^T\textbf{y}  \\
&& \textrm{s.t. } \textbf{A}\textbf{y} \le \textbf{x} 	\\
&& 0 \le \textbf{y} \le \textbf{p}^T(T-t)
\end{eqnarray*}
where $t$ and $\textbf{x}$ are inputs and $\textbf{y}$ are continuous decision variables.

This LP is a custom approximation algorithm that maximizes total revenue subject to capacity constraints. It also uses the expected demand $\textbf{p}t$ as the upper bound for solution. It is worth noting that the LP formulation is an upper bound to the DP problem (\ref{eqn_bellman1}). That is, $U_t(\textbf{x}) \ge V_t(\textbf{x})$. This problem can be solved in a fraction of a second with stanard LP solver on commodity hardware. This allows us to consider it as an ``on-demand" function call. 

In the current formulation, the LP component $U_t(\textbf{x})$ serves as a normalizer for (\ref{eqn_lp_normalizer}). First of all, it captures first-order effect, i.e., the capacity-demand trade-off in the dynamic programming problem (\ref{eqn_bellman1}) and leaves the nonlinear higher-order effect with neural network. Secondly, the LP formulation naturally captures the monotonicity of $V$ with respect to $\textbf{x}$ and $t$. Last but definitely not least, $V_t^*(\textbf{x}|\theta)$ satisfies boundary condition (\ref{eqn_boundary}) automatically, since $U_t(\textbf{x}) = 0$ when $t=T$. Using LP as a normalizer to neural networks differs the current paper from earlier LP research, e.g., in \cite{bp2003} and \cite{dfvr2003}. 

The LP ignores a few key elements of the network capacity control problem. Most notably, demand is treated as deterministic and continuous. As a result, $\Pi(U_t(\textbf{x}))$, the policy generated by LP approximation, has been shown to perform unsatisfactorily by empirical study. Refining the control policy with neural networks is the task of the next subsection.

\subsection{Bellman Error Minimization and Stratified Sampling}
Given the functional approximator $V_t^*(\textbf{x}|\theta)$, we now derive an approach to optimize it. Based on (\ref{eqn_bellman1}), let us first define the Bellman error as
\begin{equation*}
Q_t(\textbf{x}|\theta) = V^*_t(\textbf{x}) - E\left[\max_{u \in \Psi(\textbf{x})}\left\lbrace r_n u(\textbf{x},t,n) + V^*_{t+1}(\textbf{x}-\textbf{a}_nu) \right\rbrace\right] 
\end{equation*}

Combining the above equation with (\ref{eqn_lp_normalizer}) yields:
\begin{eqnarray*}
&& Q_t(\textbf{x}|\theta) \\
& = & U(\textbf{x},t)W(\textbf{x},t|\theta) -  \sum_{0 \le n \le N} p_n\bigg[\max_{u \in \Psi(\textbf{x})}\bigg( r_n u \\
&& + U_{t+1}(\textbf{x}-\textbf{a}_nu)W_{t+1}(\textbf{x}-\textbf{a}_nu|\theta) \bigg)\bigg]
\end{eqnarray*}

This leads to the following minimization problem:
\begin{equation}
\label{eqn_bellman_error}
\min \sum_{\textbf{x},t}L(Q_t(\textbf{x}|\theta))
\end{equation}
where $L(\dot)$ is the loss function. We choose to use $L^2$ norm of $Q$ as the loss in the paper, but other loss function can be used too.

Problem (\ref{eqn_bellman_error}) is difficult for a few reasons. First, $Q_t(\textbf{x})$ is nonlinear, likely non-convex, and can be locally non-differentiable. This makes optimization hard. Secondly, the sum needs to be taken on all combinations of $\textbf{x}$ and $t$, but the number of combination is practically infinite.

On the first challenge, empirical findings in deep learning literature suggest a gradient-based algorithm (such as SGD) can often reduce error to a satisfactory level, even though finding global minimum remains very difficult. Furthermore, policy $\Pi(V_t^*(\textbf{x}))$ as the opportunity cost can and (\ref{eqn_optimal_policy}) can still be a good approximation to $\Pi(V_t(\textbf{x}))$, the true optimal policy.

On the second challenge, instead of optimizing $\sum_{\textbf{x},t}L(Q_t(\textbf{x}|\theta))$, we optimize the following surrogate:

\begin{equation}
\min_{\theta} \sum_{\textbf{x},t \in \Phi}L(Q_t(\textbf{x}|\theta))
\end{equation}
where $\Phi$ is a list of $x,t$ pairs that is representative of the entire state space. To generate $\Phi$ for a specific problem case, we perform Monte-Carlo simulation using a simple first-come-first-serve control policy; see Algorithm \ref{alg_data_gen} for details.

\begin{algorithm}[tb]
   \caption{Data Generation}
   \label{alg_data_gen}
\begin{algorithmic}
   \STATE {\bfseries Input:} time horizon $T$, initial capacity $\textbf{x}(0)$,  probability vector $\textbf{p}$, and number of sample paths needed $S$
   \STATE Initialize $\Phi = \emptyset$
   \FOR {$s=0$ {\bfseries to} $S-1$}
   \STATE Initialize $\textbf{x}=\textbf{x}_0$
   \FOR{$t=0$ {\bfseries to} $T-1$}
   \STATE Draw a sample $n$ from the discrete distribution $Pr(X=n) = p_n$
   \STATE Add (\textbf{x},t) to $\Phi$
   \IF{$\textbf{x} \ge \textbf{a}_n$}
   \STATE $\textbf{x} = \textbf{x} - \textbf{a}_n$
   \ENDIF
   \ENDFOR
   \ENDFOR
   \STATE Return $\Phi$
\end{algorithmic}
\end{algorithm}

\subsection{Parameter Estimation}
Given the Bellman error $Q$ and an error function $L$, let us use $\nabla L(Q_t(\textbf{x}|\theta))$ to denote the gradient of loss with respect to neural network parameter $\theta$. We use a deep feedforward network to model $W_t(\textbf{x})$. The network has 5 hidden layers and 64 hidden units with sigmoid activation in each layer. 

For training, we split data $\Phi$ into minibatches. Each minibatch contains $B=64$ samples of $(\textbf{x}_b,t_b), b=1,...,B$. This allows us to approximate $\nabla L(Q_t(\textbf{x}|\theta))$ with the following minibatch proxy:

\begin{eqnarray*}
H(\theta) = \frac{1}{B}\sum_{b=1,...,B}\nabla L(Q_{t_b}(\textbf{x}_b|\theta))
\end{eqnarray*}

For each sample, we call an LP solver to calculate $U_{t}(\textbf{x})$ and then feed the results from LP into a gradient-based algorithm to update $\theta$; see (\ref{eqn_adamgrad}). We choose to use Adamgrad \cite{dhs2011} algorithm in Tensorflow to update the parameters, using adjustable learning rates in the range between 0.01 and 0.5.
\begin{eqnarray*}
\label{eqn_adamgrad}
\theta_{k+1} = f(\theta_k, H(\theta_k))
\end{eqnarray*}

\section{Validation}
We benchmark performance of the proposed algorithm (``DNN") using trained policy $\Pi(U_t(\textbf{x})W_t(\textbf{x}|\theta))$, with three other policies: a) the first-come-first-serve policy (``FCFS"), b) the naive LP-based policy(``LP") using policy $\Pi(U_t(\textbf{x}))$, and most importantly, c) a decomposition-based approximation algorithm (``DECOMP", see Chapter 3.4 of \cite{tvr2004}) using a sophiscated  combination of LP and single-resource dynamic programming algorithms. Different benchmarks serves different purposes. Comparison with FCFS policy helps us understand how much value there exists in network capacity control. Comparison with LP ensures that the neural network provides a better approximation to the value function than LP alone. Most importantly, comparing with DECOMP helps us understand the capability and potential of combining universal functional approximators and customized algorithms.

For benchmarking purposes, we generate four simulation cases and perform simulation experiments. For each case, we generate a unique combination of parameters, including number of resources, capacities, demand, and revenue. A summary of key parameters can be found in Table \ref{table_simulation_setting}; see supplemental material for complete details. 

\begin{table}[htbp]
\caption{BASIC SIMULATION SETTINGS}
\label{table_simulation_setting}
\vskip 0.15in
\begin{center}
\begin{small}
\begin{sc}
\begin{tabular}{lcccr}
\toprule
Case & $M$ & $N$ & $T$ & $\textbf{x}(0)$\\
\midrule
1 & 14 & 27 & 128,962 & 100 \\
2 & 14 & 39 & 47,811 & 500 \\
3 & 14 & 39 & 49,827 & 500 \\
4 & 14 & 39 & 18,684 &200 \\
\bottomrule
\end{tabular}
\end{sc}
\end{small}
\end{center}
\vskip -0.1in
\end{table}

The key performance metrics we compare is mean simulation reward (revenue). For every case, we generate 640 simulation runs (10 minibatches).  In each simulation run, a unique sequence of requests is generated and accept/reject decision is made by all policies. For any policy $\pi$, let $\mathcal{V}(\pi)$ be the mean simulation reward averaged over the 640 runs. Using DECOMP as the baseline, we calculate $\frac{\mathcal{V}(\pi)}{\mathcal{V}(DECOMP)}-1$ and show them as percentages in Table \ref{table_performance_compare}. A positive number of 1.0 means policy performs $1\%$  better than DECOMP. Raw simulation reward values are in supplement material. 
The proposed algorithm performs well compared DECOMP policy, a highly customized approximation algorithm. It also out-performs FIFO and LP policies. 

\begin{table}[htbp]
\caption{PERFORMANCE COMPARISON \\ DIFFERENCE FROM DECOMP, IN \%}
\label{table_performance_compare}
\vskip 0.15in
\begin{center}
\begin{small}
\begin{sc}
\begin{tabular}{lcccr}
\toprule
Case & DECOMP & DNN & FIFO & LP \\
\midrule
1 & - & -0.3 & -10.0 & -1.2 \\
2 & - & 0.4 & -27.1 & 0.3 \\
3 & - & -0.3 & -32.8 & -0.8 \\
4 & - & 0.3 & -31.4 & -1.5 \\
\bottomrule
\end{tabular}
\end{sc}
\end{small}
\end{center}
\vskip -0.1in
\end{table}

\section{Conclusion}
We proposed a novel approach to solve large-scale dynamic programming problems. The approach combines deep neural network, LP normalization, and stratified sampling to develop an approximate solution to the Bellman equation and a control policy. Simulation results show that this approach performs competitively against state-of-the-art deposition algorithms.

\bibliography{bellman}
\bibliographystyle{rldm2019}


\end{document}